\documentclass[authoryear,preprint,review,12pt]{elsarticle}

\usepackage{times}
\usepackage{epsfig}
\usepackage{graphicx}
\usepackage{amsmath}
\usepackage{amssymb}
\usepackage{threeparttable}
\usepackage{booktabs}
\usepackage{natbib}
\usepackage{color}
\usepackage[ruled,linesnumbered]{algorithm2e}
\usepackage{setspace}
\setcitestyle{numbers,square}

\def\ft#1{\textbf{#1}}
\def\st#1{\underline{#1}}
\def\red#1{\textcolor{black}{#1}}

\newcommand{\tabincell}[2]{\begin{tabular}{@{}#1@{}}#2\end{tabular}}

\journal{Pattern Recognition}

\begin{document}

\begin{frontmatter}

\title{Learning Generalized Visual Odometry Using Position-Aware Optical Flow and Geometric Bundle Adjustment}

\author[label1]{Yijun Cao}
\ead{yijuncaoo@gmail.com}
\author[label1,*]{Xianshi Zhang}
\ead{zhangxianshi@uestc.edu.cn}
\author[label1]{Fuya Luo}
\author[label1]{Peng Peng}
\author[label2]{Chuan Lin}
\author[label1]{Kaifu Yang}
\author[label1]{Yongjie Li}

\affiliation[label1]{organization={the MOE Key Laboratory for Neuroinformation, 
the School of Life Science and Technology, University of
Electronic Science and Technology of China},
            city={Chengdu},
            postcode={610054},
            country={China}}

\affiliation[label2]{organization={the College of Electric and Information Engineering, 
Guangxi University of Science and Technology},
            city={Liuzhou},
            postcode={545006},
            country={China}}

\affiliation[*]{organization={Corresponding authors}}

\begin{abstract}
Recent visual odometry (VO) methods incorporating geometric algorithm into deep-learning architecture
have shown outstanding performance on the challenging monocular VO task.
Despite encouraging results are shown, previous methods ignore the requirement of generalization capability
under noisy environment and various scenes. 
To address this challenging issue, 
this work first proposes a novel optical flow network (PANet).
Compared with previous methods that predict optical flow as a direct regression task,
our PANet computes optical flow by predicting it into the discrete position space with optical flow probability volume, 
and then converting it to optical flow.
Next, we improve the bundle adjustment module to fit the self-supervised training pipeline 
by \red{introducing} multiple sampling, \red{ego-motion initialization}, dynamic damping factor adjustment, and Jacobi matrix weighting. 
In addition, a novel normalized photometric loss function is advanced to improve the depth estimation accuracy.
The experiments show that the proposed system not only achieves comparable performance with other state-of-the-art
self-supervised learning-based methods on the KITTI dataset,
but also significantly improves the generalization capability compared with geometry-based, learning-based and hybrid VO systems
on the noisy KITTI and the challenging outdoor (KAIST) scenes.
\end{abstract}


\end{frontmatter}

\section{Introduction}
Simultaneous localization and mapping (SLAM) or visual odometry (VO) for estimating depth and 
relative camera poses from image sequences is a fundamental task in the fields of computer vision and robots,
involving numerous applications, \red{e.g.,} robotics navigation 
\cite{DeSouza-pami02} and augmented reality \cite{Newcombe-iccv11}. Conventional VO algorithms 
\cite{Mur-tro17, Engel-pami18} estimate camera motion via multi-view geometry, such as bundle adjustment (BA) and epipolar geometry. 
However, geometry-based VO systems may easily get failure in tracking
when illumination and texture are insufficient to establish correspondence, 
insufficient overlap exists between consecutive frames, and the scenes contains dynamic objects. 

With the strong capability of convolutional neural networks (CNNs) \cite{He-cvpr16}, 
learning-based VO models \cite{Zhou-cvpr17,Bian-nips2019} have been proposed to address these issues. 
In general, they jointly train two networks to separately predict the
depth and camera motion based on image reconstruction error.
However, deep-learning based VO approaches greatly depend on the distribution of training data and 
lack of generalization in many real situations, 
such as rapid camera movement, random noise, and cross evaluation on different datasets. 
These aspects can greatly degrade the VO performance, which has not been fully evaluated in learning-based algorithms.
In addition, learning-based VO methods \red{sometimes fail} to provide sufficient reliability and accuracy compared with those 
geometry-based methods under geometrically favorable conditions. 

\begin{figure*}[t]
    \centering
    \includegraphics[width=10cm]{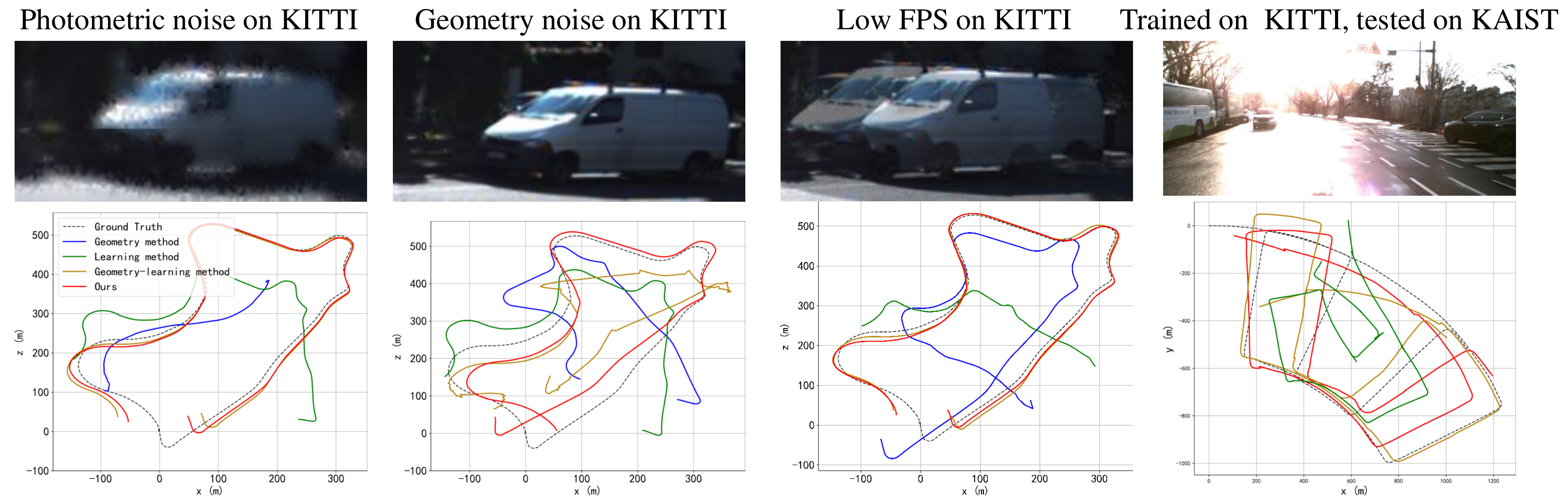}
    \caption{Comparison of generalization capability.
    All the methods, including ours, geometry method \cite{Engel-pami18}, learning method \cite{Bian-nips2019}, 
    and geometry-learning method \cite{Zhao-cvpr20}, were trained on the KITTI raw dataset, and evaluated on 
    the noisy KITTI dataset and KAIST urban dataset \cite{jeong-ijrr19}.
    }
    \label{fig:intro}
\end{figure*}

Recently, researchers have tried to combine the advantages of learning- 
and geometry-based methods to estimate VO using separate \cite{Yang-cvpr20,Zhan-icra19} or 
integrated \cite{Zhao-cvpr20} training strategies. 
These methods utilize depth and optical flow networks for predicting stable depth or correspondence information,
and use geometric module for estimating accurate ego-motion.
However, most of these schemes ignore the requirement of generalization while improving accuracy. 
Figure \ref{fig:intro} shows four common problems in VO tasks.
It can be seen that under conditions with high noise, fast motion (or relatively low frames per second, FPS) and varying scenes, 
existing VO systems have large performance fluctuations.

To avoid performance degradation and failure of tracking due to noise and domain transformation, 
this work made improvements in the following two aspects.
The first one is optical flow estimation. 
As an input to many geometric methods, such as epipolar geometry, perspective-n-point (PnP) and geometric BA \cite{gao-vslam}, 
accurate optical flow implies accurate pose estimation.
Previous optical flow models, such as PWC-Net \cite{Sun-cvpr18}, 
achieved excellent performance with CNNs. 
However, the robustness of evaluating optical flow is unclear under noisy environments.
Inspired by attention mechanism, which proves to contribute largely to models' generalization and accuracy in many applications \cite{xie2021segformer},
we proposed a new position-aware optical flow network (PANet) 
that can effectively learn the optical flow to improve robustness and accuracy.
Note that the whole model of our work is trained in self-supervised end-to-end manner.
Compared with supervised methods, which in general require expensively obtained ground truth,
self-supervised manner can train models only on monocular videos.

The second aspect we focused on improving is pose estimation.
Compared with using epipolar geometry to compute camera motion \cite{Zhao-cvpr20,Zhan-icra19}, 
the BA algorithm can better avoid the disruption of outliers due to the Huber function \cite{Engel-pami18}.
BA can use different metrics to optimize the poses, 
including geometric metric
\cite{Mur-tro17}, 
photometric metric
\cite{Engel-pami18}, 
and feature metric
\cite{Tang-iclr19}.
In this work, we select geometric metric, because 
1) compared with photometric metric, which only uses the RGB feature, computing the correspondence from optical flow can use 
the powerful CNN and achieve better performance.
2) feature metric \red{based} methods are very difficult to converge, especially in self-supervised manner.
In contrast, we use geometric metric based BA as our pose estimation model.
In our experiment, the training will fail when 
directly embedding BA with conventional Levenberg-Marquardt (LM) algorithm \cite{gao-vslam} into deep learning pipeline.
\red{This is because} BA requires suitable depth for initialization, 
whereas the depth network \red{can} not provide accurate depth at the early training stage.
To help the network converge, we improve the BA module with many tricks, 
including multiple sampling, \red{ego-motion initialization}, dynamic damping factor adjustment, and Jacobi matrix weighting.

In summary, the main contributions of this paper lie in two aspects: 
\begin{itemize}
    \item[(1)] A novel position-aware flow network (PANet) is proposed,
    which converts the regression estimation to locally position-aware model with optical flow probability volumes.
    It barely increases the number of parameters while improving performance in terms of
    accuracy and generalization capability.
    \item[(2)] \red{We embed} the geometric BA module with the Levenberg-Marquardt (LM) algorithm \cite{gao-vslam} 
    into a joint depth and optical flow learning system and train them together in self-supervised manner.
    Several strategies are employed:
    i) using multiple filtering to ensure the stability of the input pixels; 
    ii) dynamically adjusting the damping factor $\lambda$ according to the error of each training batch, 
    iii) assigning different weights to the Jacobi \red{matrices} of pose and inverse depth; and
    iv) initializing with PnP. 
    These strategies help achieve a better convergence of the BA module throughout the training process.
\end{itemize}

Experiments show that the proposed system not only achieves state-of-the-art performance on the KITTI dataset,
but also significantly improves the generalization capability across many challenging scenes 
in a self-supervised learning manner, as indicated in Fig. \ref{fig:intro}.

\begin{figure*}[t]
    \centering
    \includegraphics[width=10cm]{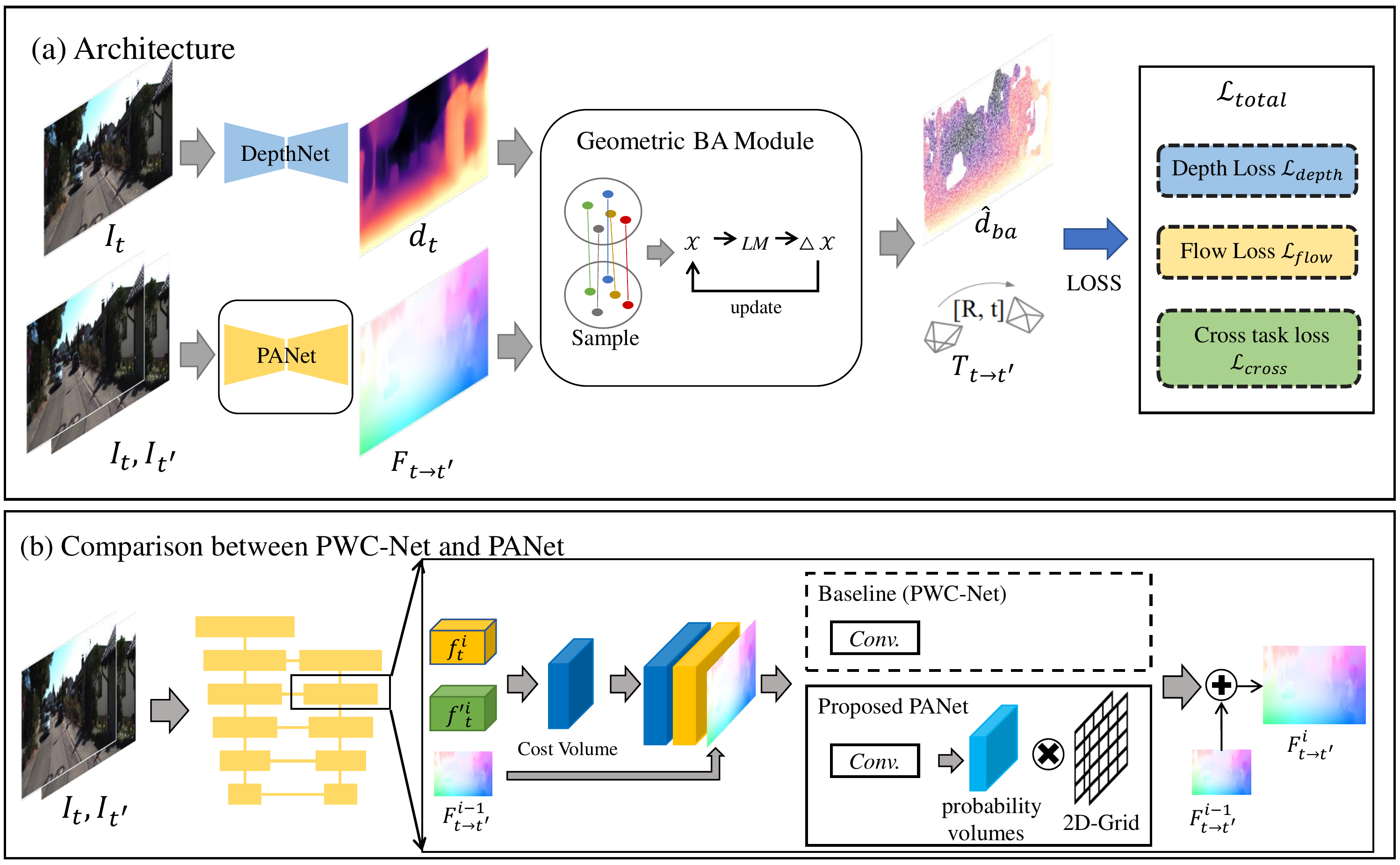}
    \caption{(a) Overview of proposed \red{architecture}.
    The Depth-Net and PANet estimate the inverse depth $d_t$ and optical flow $F_{t \rightarrow  t'}$, respectively.
    The geometric BA module receives the predicted dense $d_t$ and $F_{t \rightarrow  t'}$, 
    samples them through several strategies, and finally outputs sparse inverse depth $\hat{d}_{ba}$
    and ego-motion $T_{t \rightarrow t'}$ by optimizing the re-projection error using the LM algorithm.
    \red{
    (b) Comparison between PWC-Net and PANet. Both of them use a encoder-decoder architecture.
    The feature map $f$ from the first image and warped from the second image $f'$
    are fed to compute the cost volume. 
    Then, cost volume, $f$, and the optical flow predicted by the last estimator $F_{t \rightarrow  t'}^{i-1}$ are concatenated.
    PWC-Net \cite{Sun-cvpr18} predicts flow using several convolutional layers. In contrast, the proposed PANet first computes 
    probability volume (PV), and then converts it to optical flow.
    }
    }
    \label{fig:overview}
 \end{figure*}

\section{Related Work}
\textbf{Learning based monocular depth estimation.} 
With the development of CNNs, a variety of models have been proposed to learn monocular depth 
in a supervised manner \cite{Eigen-iccv15}. However, these methods require labeled ground truth, which are 
expensive to obtain in natural environments. More recent works have begun to approach the problem in a self-supervised or unsupervised way. 
A pioneering work is SfMLearner \cite{Zhou-cvpr17}, which learns depth and ego-motion jointly by minimizing photometric loss in an unsupervised manner.
This pipeline has inspired a large amount of follow-up works. To deal with moving objects breaking the assumption of static scenes, 
many works \cite{Yin-cvpr18, Luo-pami20} use consistency of forward-backward optical \cite{Zhao-cvpr20}, depth-optical \cite{Zou-eccv18}, 
or depth-depth \cite{Bian-nips2019} flow to mask dynamic objects.
Several methods develop new frameworks by changing training strategies and adding supplementary constraints \cite{Godard-iccv19}
and collaborative competition \cite{Ranjan-cvpr19}.
More recently, several researchers \cite{Zhao-cvpr20,Yang-cvpr20} have tried to combine a geometric algorithm into the deep learning architecture, and obtained 
better depth and VO estimations by training with only two frames in a video sequence. Different from those methods, the proposed method embeds geometric BA in the entire system, which 
predicts ego-motion and depth simultaneously and improves both accuracy and generalization.

\textbf{Optical flow estimation.} 
Optical flow techniques attempt to get apparent movement of brightness patterns between continuous images. 
Classical methods infer optical flow for a pair of images by minimizing photometric consistency and smoothness \cite{Lucas-ijcai81}. 
Recent approaches, such as PWC-Net \cite{Sun-cvpr18}, regard optical flow estimation as a supervised learning problem with CNN
\red{and can be better applied to other optical flow-based tasks \cite{zhang2020learning}. }
Although these methods produce outstanding performance by incorporating useful components,
supervised methods require accurate optical flow labels, which is laborious, and requires approaches as unusual as manually painting the scenes with textured fluorescent paint 
and imaging them under ultraviolet light.
In contrast, unsupervised methods overcome the need for labels by optimizing photometric consistency and local flow smoothness. Some recent approaches significantly 
improve the results by omitting occluded regions using forward-backward consistency checks \cite{Sundaram-eccv10} or range-map filtration \cite{Wang-cvpr18}.
Other extensions introduce edge-aware smoothness \cite{Wang-cvpr18}, data distillation \cite{Sundaram-eccv10}, 
and co-training optical flow with depth and ego-motion models \cite{Luo-pami20,Zhao-cvpr20}. 
Differently, the proposed optical flow network is built on PWC-Net \cite{Sun-cvpr18} training with self-supervised manner \cite{Zhao-cvpr20}
and further improves the performance.

\textbf{Visual odometry.} VO is a long-standing problem that estimates the ego-motion incrementally using visual input. 
Benefiting from the theory of multi-view geometry, a geometry-based VO system usually consists of two steps. 
First, the raw camera measurements are processed to generate a photometric \cite{Newcombe-iccv11,Engel-pami18} or feature \cite{Mur-tro17} representation. 
Second, the representation is used to estimate depth and ego-motion using geometry methods 
(e.g., epipolar geometry \cite{Mur-tro17}, \red{and structural constraints \cite{company2022msc}) }
and local optimization (e.g., photometric BA \cite{Engel-pami18}). 
Recently, many researchers tried to solve the VO problem using CNNs in a supervised or self-supervised manner. 
The supervised methods minimize the distance between predicted values (depth and ego-motion) and corresponding ground truth by using 
such strategies as a recurrent neural network (RNN) \cite{Wang-icra17}, or feature-metric BA \cite{Tang-iclr19}.
In contrast, to avoid the need for annotated data, self-supervised VO was developed using structure-from-motion (SfM) pipeline. These methods 
accept continuous image input and infer VO via a CNN \cite{Zhou-cvpr17,Bian-nips2019} or LSTM \cite{Zou-eccv20,Li-cvpr20}.
To combine the advantages of geometry-based and deep-learning methods, several works \cite{Zhan-icra19,Yang-cvpr20} 
separately learned the various components (e.g., optical flow, depth, and VO) of the entire system; others \cite{Zhao-cvpr20} tried to combine them and train them jointly.
In the present work, the joint learning framework is further improved with the proposed PANet and a geometric BA module.

\section{Method}
\subsection{Overview}
\label{sec:Method-Overview}
As shown in Fig. \ref{fig:overview}(a), the overall architecture of the proposed system includes three components: 
Depth-Net \cite{Godard-iccv19}, the proposed PANet, and a geometric BA module, 
which are unified into a framework working in a self-supervised way.
Given a pair of consecutive frames $I_t$ and $I_{t'}$, 
the Depth-Net and proposed PANet output inverse depth $d_t$ and optical flow $F_{t \rightarrow t'}$, respectively.
The BA module receives $d_t$ and $F_{t \rightarrow  t'}$, and outputs sparse inverse depth and 
ego-motion $T_{t \rightarrow t'}\in \mathcal{S} \mathcal{E} (3)$. 

In the following, Section \ref{sec:Method-depth} briefly describes the depth network used in this paper,
and then we describe the network architecture of the proposed PANet (Section \ref{sec:Method-Flow}).
Next, we demonstrate the geometric BA module for suiting deep learning architectures (Section \ref{sec:Method-BA}).
Finally, a detailed description of the loss functions and training strategy are presented in Section \ref{sec:Method-TI}.

\begin{figure}[t]
    \centering
    \includegraphics[width=6cm]{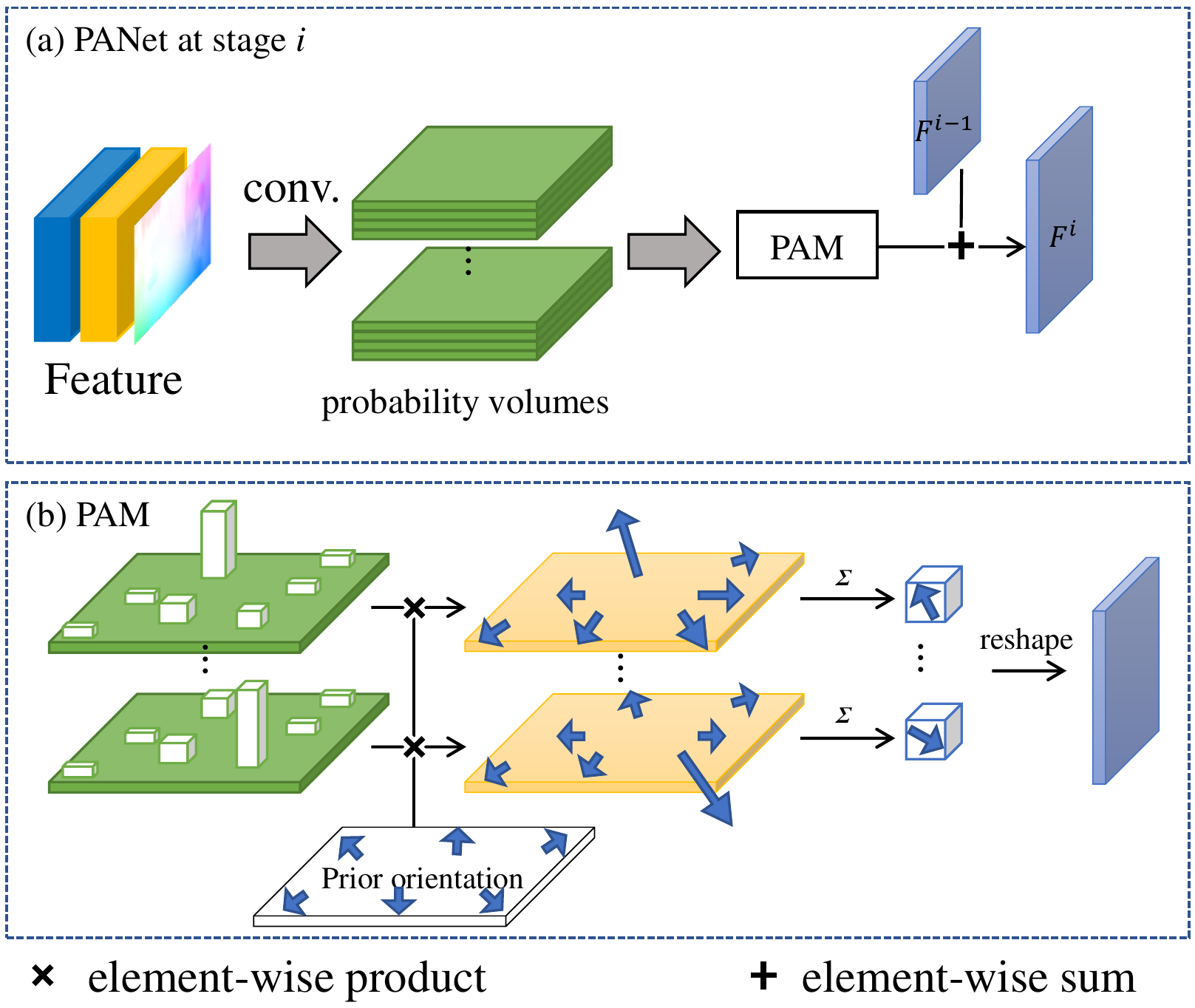}
    \caption{
    (a) The proposed PANet at the decoding stage $i$. 
    The input consists of the features of the first image, the cost volume, and the up-sampled optical flow at the previous stage $i-1$.
    PANet first predicts optical flow probability volume
    and then computes the optical flow at stage $i$ \red{by concatenating the} PAM outputs (the local optical flow in stage $i$) 
    and the up-sampled optical flow at the previous stage $i-1$.
    (b) For each channel, the PAM first computes the magnitude of optical flow 
    with regard to a priori local orientation table along the $x$ and $y$ axes
    The magnitude then is averaged \red{at} all orientations to obtain the optical flow at each pixel of 2-D space $H \times W$.
    }
    \label{fig:flow}
 \end{figure}

\subsection{Depth network} 
\label{sec:Method-depth}
For the depth network, the same architecture as \cite{Godard-iccv19} was used, which adopts 
encoder-decoder design with skip connections and 5 scales side outputs.
The encoder is ResNet18 \cite{He-cvpr16} without full connection layers;
at each scale, the decoder consists of 2 convolutional layers with the kernel size of 3 and 
1 convolutional layers with the kernel size of 1.
The activation function in decoder is ELU \cite{DBLP:journals/corr/ClevertUH15}.
Mirrored exponential disparity (MED) \cite{Gonzalez-nips20} was substituted
for a conventional Sigmoid function \cite{Godard-iccv19} as the output activation process.

\subsection{PANet}
\label{sec:Method-Flow}
The proposed PANet 
is built upon PWC-Net \cite{Sun-cvpr18}, which is composed of two compact components:
a pyramidal feature extractor and several optical flow estimators at each resolution (Figure \ref{fig:overview}(b)). 
The feature extractor is a six-stage convolutional architecture, and each stage includes three $3\times3$ convolution layers activated by LeakyReLU.
From coarse to fine, the optical flow estimators receive concatenated features that consist of 
cost volume, the feature from the first frame $I_t$, and the optical flow $F_{t \rightarrow t'}^{i-1}$ from the $i-1$-th stage. 
Given the feature map and warping feature, the cost volume layer computes the matching costs for associating 
a pixel with its corresponding pixels at the next frame in a local region.

The optical flow estimator \cite{Sun-cvpr18} \red{at} each stage of different resolution (PWC-Net in Figure \ref{fig:overview}(b))
predicts optical flow straightforward \red{simply via several convolutional layers}, 
while the proposed PANet (Figure \ref{fig:flow}(a)) introduces a priori of position of the optical flow, 
which is defined as position-aware optical in this work.
\red{To predict the correspondence of pixels using optical flow from frame} $t-1$ to $t$,
previous works, such as PWC-Net \cite{Sun-cvpr18}, regress it as a 2-dimensional(D) vector, i.e., the offset of $x$ and $y$ axes, in a scope of $[-k, k]$.
Inspired by attention mechanism, which has proved its promising capacity of generalization and accuracy in many applications 
\cite{xie2021segformer}, 
we believe that a local 2D grid containing all possible optical flow directions can be constructed as a priori,
and the probability of each optical flow in a local can be predicted by the network.
This priori position is about the possible optical flow of a pixel from the frame $I_t$ to the frame $I_{t'}$,
i.e., the offset value, defined as two 2-D grids along the $x$ and $y$ axes:
\begin{equation}
    G_x = 
    \begin{bmatrix}
        -k & \cdots & k \\
        \vdots & \ddots & \vdots \\
        -k & \cdots  & k
    \end{bmatrix}, 
    G_y = 
    \begin{bmatrix}
        k & \cdots & k \\
        \vdots & \ddots & \vdots \\
        -k & \cdots  & -k
    \end{bmatrix},
\end{equation}
where $k$ is set to 4, the same as previous work \cite{Sun-cvpr18}; 
$G_x$ and $G_y$ denote the possible optical flow values of a local region along the x and y directions, respectively.
With a priori coordinates, the model predicts a 3-D optical flow probability volume 
(OPV $\in \mathbb{R}^{\mathcal{K} \times \mathcal{K}\times H W}$), which can be viewed as a weight matrix, 
storing the probability of the optical flow to fall in the area with a size of $\mathcal{K}\times \mathcal{K}$ pixels,
where $\mathcal{K}=2k+1$. 
Note that the PWC-Net outputs a 2-D vector at each spatial position $\mathbb{R}^{2 \times H W}$, 
and we estimate $\mathcal{K} \times \mathcal{K}$ matrix which indicates the probability in local position.
Next, as shown in the Figure \ref{fig:flow}(b),
the optical flow residuals at the $i$-th stage can be expressed as 
a weighted average of OPV over the predefined 2D position matrices $G_x$ and $G_y$, 
formulated as 
\begin{equation}
   \begin{split}
      &F_x(i, j) = \frac{1}{\mathcal{K}^2} \sum_{u=-k}^{k}\sum_{v=-k}^{k} OPV(u, v, i, j)  G_x(u, v),\\
      &F_y(i, j) = \frac{1}{\mathcal{K}^2} \sum_{u=-k}^{k}\sum_{v=-k}^{k} OPV(u, v, i, j)  G_y(u, v),
   \end{split}
\end{equation}
where $F_x(i, j)$ and $F_y(i, j)$ are the components of the optical flow residuals along the $x$ and $y$ axis, respectively.
The final optical flow at the $i$-th stage is expressed as the sum of the residuals and the optical flow at the previous stage.
The `position-aware" means that the process requires the network to learn an awareness matrix that applies weighting on the predefined optical flow positions.

Optical flow probability volumes can be understood as a way of optical flow discretization. 
Similar to disparity discretization \cite{Gonzalez-nips20}, 
it might be reasonable to use quantization optical flow position because the range of optical flow is limited in a local area.
Optical flow at each stage only has fixed value in a 2-D grid $G_x$ and $G_y$, 
so each convolutional kernel of proposed PANet only need to predict a probability of one position.
Combining the two analyses mentioned above, we believe that the position-aware mechanism is effective,
and our experiments will show that the proposed PANet outperforms PWC-Net based self-supervised methods \cite{Zhao-cvpr20,Luo-pami20}
in terms of both accuracy and generalization 
(Table \ref{table:flow}).

\subsection{Geometric BA module}
\label{sec:Method-BA}
\textbf{Correspondence and Sample.} Optical flow provides dense correspondence for each pixel. 
However, not all correspondences can be used correctly 
due to the errors of optical flow and noise caused by moving objects.
Therefore, it is very important to filter out the noise and unnecessary points before BA.
Inspired by previous studies \cite{Zhao-cvpr20, Zhan-icra19}, 
screening is done at three stages in our model to ensure the reliability of the pixels' correspondence. 
Given a forward optical flow $F_{t \rightarrow  t'}$ and backward optical flows $F_{t' \rightarrow  t}$,
the first filtering condition is flow consistency mask $M_c$, 
which is composed of a forward-backward consistency mask \cite{Liu-cvpr19} and occlusion mask \cite{Wang-cvpr18}. 
The forward-backward consistency mask is benefit to get stable correspondences
by selecting the part with small difference of forward and backward optical flow.
Occlusion mask helps filter out the occluded areas by checking whether pixels are filled by
warped backward optical flow.
After obtaining $M_c$, good 2D-to-2D correspondences can be chosen according to the top 20\% consistency from 
$S_o = M_c /{|-F_{t \rightarrow  t'}-F_{t' \rightarrow  t}|} $.
The second key operation is the removal of outlier pixels. 
After computing a fundamental matrix with 3,000 best correspondences from the first step, 
the inlier score is computed as $S_r=\mathbf{1}(D_e < 0.5) / (1.0+D_e)$, 
where $D_e$ is the distance map representing the distance from pixels to their corresponding epipolar line, and
$\mathbf{1}(\cdot )$ denotes indicator function.
The last process is to filter out the matches that have minimal ray angles or invalid re-projection \cite{Zhao-cvpr20}, 
and the filter is expressed as a mask, $M_a$.
Specifically, we compute the cosine value of angle between the two rays $L_1$ and $L_2$ in 3-D space 
which is computed by correspondence pixels. Then, the mask $M_a$ filters out the regions with small cosine value.
To improve computational efficiency, the best 3,000 matches from $M_a S_r S_o$ are randomly sampled.

\textbf{Geometric BA.} 
After filtering and sampling, several inverse depths $\hat{d}$ can be obtained and then fed into the geometric BA module.
The BA \cite{Agarwal-eccv10} jointly optimizes camera poses $T$ and 
3D scene coordinates by minimizing the re-projection error, 
and has been regarded as the gold standard for SfM in the last two decades. 
To embed it into the proposed training pipeline, the BA algorithm was simplified into two-view form and 
only the camera poses $T$ and inverse depth $d$ of several points were optimized.
The objective function is formulated as 
\begin{equation}
   \mathcal{X}  = \mathop{\arg\min}_{T, d} \sum_{i=1}^{N}\nolimits \omega |\mathcal{E}_i|, 
   \label{eq:ba}
\end{equation} 
where $\mathcal{X}=\{T, d_i | i=1,2,...,N\}$. $N$ is the number of points to be optimized.
$\mathcal{E}$ is the error of geometric distance, defined as
$\mathcal{E}_i = p_i - P'_{depth}(P_{flow}(p_i))$, where the function $P'_{depth}(p_i)$ is the inverse transformation of $P_{depth}(p_i)$,
and $P_{depth}(p_i)$ as well as $P_{flow}(p_i)$ are 
the camera perspective projection and photometric correspondence, respectively, 
which are formulated in Eq. \ref{eq:1}.
The weight $\omega$ is computed with Huber loss \cite{Engel-pami18}. 
The symbol $d_i$ is the inverse depth of the $i$th pixel.

The general strategy to minimize Eq. \ref{eq:ba} is to use the LM algorithm. 
At each iteration, the LM algorithm obtains an optimal update $\Delta \mathcal{X}^{*}$ 
to the solution by minimizing:
\begin{equation}
    \Delta \mathcal{X}^{*}  = \arg\min || J(\mathcal{X}) \Delta \mathcal{X} + E(\mathcal{X}) ||^2 + \lambda || D(\mathcal{X}) \Delta\mathcal{X} ||^2, 
    \label{eq:LM}
\end{equation} 
where $E(\mathcal{X})=[\mathcal{E}_1, \mathcal{E}_2, ..., \mathcal{E}_N]$, 
$J(\mathcal{X})=[J_{T}, J_d]$ is the Jacobian matrix of $E(\mathcal{X})$ with respect to camera poses $T$ and inverse depth $d$,
$D(\mathcal{X})$ is a non-negative diagonal matrix, typically the square root of the diagonal of the
approximated Hessian $J(\mathcal{X})^{\intercal}J(\mathcal{X})$. 
The non-negative value $\lambda$ controls the regularization strength.

However, a standard LM algorithm is not suitable for end-to-end self-supervised training due to its two main problems. 
The first one is the fixed parameter initialization on damping factor $\lambda$ and ego-motion $T$. 
On the premise of ensuring as few iterations as possible, a ego-motion $T$ with large error or a small $\lambda$
will make the algorithm difficult to converge to the minimum, 
while a large $\lambda$ will lead to a local optimal solution. 
To solve this problem, PnP algorithm \cite{Ke-cvpr17} was targeted to do ego-motion $T$ initialization.
Compared with the initial attempt with an identity matrix, 
the errors of both using PnP and an identity matrix are high at the early stage of model training,
while at the later stages of training, PnP can better approximate the optimal solution.
The following equation is also used to initialize $\lambda$, 
\begin{equation}
   \lambda =  \lambda_{min} + (\lambda_{max} - \lambda_{min}) exp({-\frac{\frac{1}{N} \sum_{i=1}^{N}\nolimits \omega |\mathcal{E}_i|}{\sigma}}),
   \label{eq:damp_factor}
\end{equation}
where the minimum $\lambda_{min}=1$, maximum $\lambda_{max}=10^{4}$ and $\sigma=5$ are empirically set for training. 
The damping factor $\lambda$ controls whether BA prefers to use first- or second-order optimization.
When the error is large, Eq. \ref{eq:damp_factor} makes $\lambda$ to decrease so that the LM algorithm tends to converge quickly with the second-order method.
In contrast, when the error is small, the equation makes $\lambda$ to increase 
so that the LM algorithm tends to increase the convergence accuracy with the first-order method.

The second issue is an empirical discovery that the numerical ratio of the Jacobi matrix of $T$ and $d$
will greatly affect the training result of the entire model, especially Depth-Net.
In the training phase, if the depth Jacobi matrix $J_d$ is given a small weight ($J(\mathcal{X})=[J_{T}, w_d * J_d]$, where $w_d<1$), 
then the training accuracy of Depth-Net will be essentially improved (see Table \ref{table:depth}), 
and conversely, in the testing phase, if the small weight of the depth Jacobi matrix is still maintained, 
the testing error of the ego-motion will be particularly large (see Table \ref{table:vo}).
It is conjectured that this phenomenon is caused by the inconsistent range of values of the depth and pose Jacobi matrixes,
and smaller values will give more accurate results in the joint optimization using Schur-Complement \cite{gao-vslam}.
Based on the above finding, a small weight $w_d=0.1$ is set for the depth Jacobi matrix in the training phase, 
and it is reset to $w_d=1.0$ during the inference phase.

\red{The inference process of the model, i.e., one feedforward propagation process in training, 
is shown in Algorithm \ref{alg:inference}.}

\resizebox{\linewidth}{!}{
\begin{algorithm}[H]
    \footnotesize
    \setstretch{0.8}
    
    \label{alg:inference}
    
    \caption{\red{Model inference with monocular videos}}
    \LinesNumbered
    \KwIn{a pair of consecutive frames $I_t$ and $I_{t'}$}
    \KwOut{Depth {$D_{t}$}, ego-motion {$T_{t \rightarrow t'}$}, and optical flow {$F_{t \rightarrow  t'}$}}

    $d_{t}$ $\leftarrow$ Depth-Net($I_t$) \tcp*{inverse depth}
    $F_{t \rightarrow  t'}$ $\leftarrow$ PANet($I_t$, $I_{t'}$) \tcp*{forward optical flow} 
    $F_{t' \rightarrow  t}$ $\leftarrow$ PANet($I_{t'}$, $I_{t}$) \tcp*{backward optical flow} 
    Compute $M_a, S_r, S_o$ from $F_{t \rightarrow  t'}$ and $F_{t' \rightarrow  t}$ \tcp*{compute masks}
    Sample $N$ pixels $d_{t}^{i=1...N}, F_{t \rightarrow  t'}^{i=1...N}$ according to $M_a \times S_r \times S_o$ \;

    $T_0 \leftarrow$ P3P($F_{t \rightarrow  t'}, d_{t}$) \tcp*{init pose} 
    Initialize $\lambda$ using Eq. \ref{eq:damp_factor} \;
    $\mathcal{X}_0 \leftarrow \{T_0, d_{t}^{i=1...N}\} $ \;
    \For(){$1\leqslant j \leqslant K$}{
        Compute geometry-metric error $\mathcal{E}$ using $\mathcal{X}_j$ and $F_{t \rightarrow  t'}^{i=1...N}$ \;
        Jacobi matrix $J(\mathcal{X}) \leftarrow [J_{T}, w_d * J_d]$ \;
        Hessian matrix $H(\mathcal{X}) \leftarrow J(\mathcal{X})^{T} J(\mathcal{X})$ \;
        Diagonal matrix $D(\mathcal{X}) \leftarrow Diag(H(\mathcal{X}))$ \;
        Solve $\Delta \mathcal{X} \leftarrow (H(\mathcal{X}) + \lambda D(\mathcal{X}))^{-1}J(\mathcal{X})^{T}\mathcal{E}$  \tcp*{LM step}
        $\mathcal{X}_j \leftarrow Updata(\Delta \mathcal{X}, \mathcal{X}_{j-1})$ \;
    }
    $ T_{t \rightarrow t'}, d_{t} \leftarrow \mathcal{X}_K$ \;
    $D_{t} \leftarrow 1 / d_{t}$ \;
    
\end{algorithm}
}
\subsection{Training the system}
\label{sec:Method-TI}
\textbf{Self-supervised learning pipeline.}
The method for finding corresponding pixels with regard to depth and optical flow is introduced first.
For a pixel $p_t$ in $I_t$, the corresponding pixel $p_{t'}$ in $I_{t'}$ can be found either
through camera perspective projection $P_{depth}(p_t)$ or the photometric correspondence $P_{flow}(p_t)$,
which are consistent for static scenes. 
Formally, the two relationships can be written as
\begin{equation}
   \begin{split}
   &P_{depth}(p_t) = KT_{t \rightarrow t'}D_{t}(p_{t})K^{-1}p_{t}, \\
   &P_{flow}(p_t) = p_{t} + F_{t \rightarrow t'}(p_{t}),
   \end{split}
   \label{eq:1}
\end{equation}
where $K$ and $D_{t}(p_{t}) = 1/d_{t}(p_{t})$ denote the camera intrinsic and the depth in $p_{t}$, respectively.

After computing the corresponding $p_t$ and $p_{t'}$, the \red{synthetic image $I'_{t}$ can be warped using $I_{t'}$}.
Then, self-supervised training of both depth and optical flow
is realized by minimizing the photometric error between $I_t(p_t)$ and the synthetic image $I'_t(p_t)$:
\begin{equation}
   \mathcal{L}_{self}^{\mathbb{S}}(W) = \sum\nolimits_{p_t} W(p_t)r(I_t(p_t), I'_t(p_t)),
\end{equation}
where $\mathbb{S} \in \{F_{t \rightarrow t'}, F_{t' \rightarrow t}, d_t, d_{t'}\}$ indicates the way in which the images are synthesized to train the self-supervised loss $\mathcal{L}_{self}^{\mathbb{S}}(W)$.
$W(p_t)$ is a weight indicating credibility in $p_t$. The function $r(I_t(p_t), I'_t(p_t))$ is the metric
between the target image $I_t(p_t)$ and synthetic image $I'_t(p_t)$, and is usually defined as L1 + SSIM \cite{Zhou-tip04}:
\begin{equation}
   r(I_t, I'_t) = \frac{\alpha}{2}(1-SSIM(I_t, I'_t)) + \alpha|I_t-I'_t|,
   \label{eq:l1_ssim}
\end{equation}
where $\alpha=0.85$ is set for the training process. $SSIM(\cdot)$ is the function of computing structural similarity \cite{Zhou-tip04}.

Pixel-level color matching alone is unstable and ambiguous. Therefore, an
edge-aware smoothness term is often applied for regularization \cite{Luo-pami20}:
\begin{equation}
   \mathcal{L}_{s}(\mathbf{V}, \beta, k) = \sum_{p_t} \sum_{\mathbf{r}\in x,y} |\frac{\partial^k\mathbf{V}(p_t)}{\partial \mathbf{r}^k}| e^{-\beta|\frac{\partial I_t(p_t)}{\partial \mathbf{r}}|},
   \label{eq:smoothness} 
\end{equation}
where $\mathbf{V}$ represents the type of input and $\beta$ is the order of the smoothness gradient.

\textbf{Training strategies and losses.}
In practice, it is found that jointly training all networks does not generate reasonable outputs \cite{Luo-pami20,Zhao-cvpr20}.
The possible reasons include i) random initialization of different modules (Depth-Net and PANet) 
can lead to difficulties in converging the network to the same objective function;
ii) it is difficult to train complex networks effectively with self-supervised cost functions in the presence of multiple task objectives.
According to previous work \cite{Luo-pami20,Zhao-cvpr20}, the entire training schedule consists of three stages:
(1) stage I: self-supervised training of optical flow, (2) stage II: self-supervised training of depth with fixing optical flow network weights
and (3) stage III: fine-tuning of optical flow with fixing depth network weights using cross-task loss $\mathcal{L}_{cross}$ \cite{Zou-eccv18}.

First, the proposed PANet is trained in a self-supervised manner, and the loss function $\mathcal{L}_{flow}$ consists of 
photometric error and edge-aware smoothness:
\begin{equation}
   \mathcal{L}_{flow} = \frac{1}{2} \sum_{\mathbf{F} \in \{F_{t \rightarrow t'}, F_{t' \rightarrow t}\}} \mathcal{L}_{self}^{\mathbf{F}}(M_c) + \gamma_{fs}\mathcal{L}_{s}(\mathbf{F}, 10, 2),
\end{equation}
where $M_c$ is the flow consistency mask mentioned in Section \ref{sec:Method-BA}.

Second, the PANet parameters are fixed and the Depth-Net is trained using $\mathcal{L}_{depth}$, 
which consists of normalized photometric error, geometric consistency \cite{Bian-nips2019}, 
edge-aware smoothness, and point-wise depth loss:
\begin{equation}
   \begin{split}
    \mathcal{L}_{depth} = \frac{1}{2} \sum\nolimits_{\mathbf{d} \in \{d_t, d_{t'}\}} \gamma_{drp}\mathcal{L}_{rp}^{\mathbf{d}} 
    + \gamma_{ds}\mathcal{L}_{s}(\mathbf{d}, 2, 1) + \gamma_{dc}\mathcal{L}_{dc}^{\mathbf{d}} + \gamma_{dba}\mathcal{L}_{ba}^{\mathbf{d}},
   \end{split}
\end{equation}
where geometric consistency $\mathcal{L}_{dc}=\frac{1}{N} \sum_{p_t} \hat{d}_{t\rightarrow t'}(p_{t}) - \hat{d}_{t'}(p_{t})$
is defined in SC-SfMLearner \cite{Bian-nips2019}, where
$\hat{d}_{t\rightarrow t'}$ is the synthesized depth map by warping $\hat{d}$ \red{at} $t$ to $t'$,
\red{
and $N$ is the sum of $M_d$. $M_d$ is the validation mask of depth, which 
is the product of the auto-mask \cite{Godard-iccv19}, depth projection mask \cite{Godard-iccv19}, 
and inlier score $S_r$ mentioned in Section \ref{sec:Method-BA}. 
The auto-mask \cite{Godard-iccv19} is used to filter out the pixels which do not change appearance from one
frame to the next in the sequence.
It is clear that such mask can make the network to ignore the objects which move at the same
velocity as the camera, and even to ignore the whole frames in monocular videos when the camera stops moving.
Depth projection mask \cite{Godard-iccv19} is used to filter out those pixels that are outside the image boundaries after camera movement.
}

We experimentally found that the conventional photometric error performs poorly in dark light, 
which may be due to the small weighting caused by the small pixel value in the dark region. 
Therefore, a normalized photometric error function is proposed, which is written as 
\begin{equation}
   \mathcal{L}_{rp}^{\mathbf{d}} = \frac{1}{N} \sum_{p_t}\nolimits M_d(p_t)\frac{r(I_t, I'_t)(p_t)}{I_t(p_t)},
\end{equation}
where $N$ is the sum of $M_d$; 
function $r(I_t, I'_t)$ is defined in Eq. \ref{eq:l1_ssim}. 

The point-wise depth loss defines the difference between the predicted inverse depth after sampling $\hat{d}(p_t)$ 
and the inverse depth $\hat{d}_{ba}(p_t)$ outputted by BA module, which is formulated as
\begin{equation}
   \mathcal{L}_{ba} = \frac{1}{N} \sum_{p_t}\nolimits |\hat{d}(p_t) - \hat{d}_{ba}(p_t)|.
\end{equation}

At the last training stage, the Depth-Net is fixed and the PANet is fine-tuned using cross-task loss $\mathcal{L}_{cross}$ \cite{Zou-eccv18}:
\begin{equation}
   \mathcal{L}_{total} = \gamma_{f} \mathcal{L}_{flow} + \gamma_c  \mathcal{L}_{cross},
\end{equation}
where $\mathcal{L}_{cross}$ is used to minimize the difference between $P_{depth}(p_t)$ and $P_{flow}(p_t)$. 
The two flow fields should be consistent with each other for non-occluded and static regions. 
Minimizing the discrepancy between the two flow fields can simultaneously update the depth and flow models:
\begin{equation}
   \mathcal{L}_{cross} = \frac{1}{N} \sum_{p_t}\nolimits S_r(p_t) |P_{depth}(p_t) - P_{flow}(p_t)|.
\end{equation}
Note that we fix the Depth-Net at the third stage unlike some joint learning algorithms \cite{Zhao-cvpr20, Zou-eccv18}
due that the use of cross-task loss $\mathcal{L}_{cross}$ assumes that 
the scene is static or that the optical flow and depth have a similar dynamic target mask, which is very rare in real traffic scenarios.

\begin{table*}[ht]
    \footnotesize
    \setstretch{0.8}
    
    \centering
    \caption{Quantitative results. Comparison of the proposed method with existing
    methods without post-processing on the KITTI 2015 \cite{Geiger-ijrr13} using the Eigen split \cite{Eigen-nips14}. 
    All the methods were trained in a self-supervised manner with monocular data. 
    Bolded and underlined numbers is best and second-best metrics.}
    \label{table:depth}
    \resizebox{\linewidth}{!}{
    \begin{tabular}{lc*3{c}*3{c}}
       \hline
       \multicolumn{1}{c}{}& \multicolumn{4}{c}{Error}& \multicolumn{3}{c}{Accuracy} \\
       Method& AbsRel$\downarrow$& SqRel$\downarrow$& RMS$\downarrow$& RMSlog$\downarrow$& $<1.25\uparrow$& $<1.25^2\uparrow$& $<1.25^3\uparrow$\\
       \hline
       SfMLearner \cite{Zhou-cvpr17} &           0.183&      1.595&      6.709&      0.270&      0.734&      0.902&      0.959\\
       CC \cite{Ranjan-cvpr19} &                 0.140&      1.070&      5.326&      0.217&      0.826&      0.941&      0.975\\
       EPC++ \cite{Luo-pami20} &                 0.141&      1.029&      5.350&      0.216&      0.816&      0.941&      0.976\\
       SC-SfMLearner \cite{Bian-nips2019} &      0.137&      1.089&      5.439&      0.217&      0.830&      0.942&      0.975\\
       Monodepth2 \cite{Godard-iccv19} &         0.115&      0.882&      4.701&      0.190& \st{0.879}&      0.961&      0.982\\
       Zhao et al. \cite{Zhao-cvpr20} &     \st{0.113}& \ft{0.704}&      4.581& \st{0.184}&      0.871&      0.961& \st{0.984}\\
       Sun et al. \cite{sun-tnnls21} &      \ft{0.110}&      0.791&      4.557& \st{0.184}& \ft{0.887}& \ft{0.964}&      0.983\\
       \hline
       Ours (baseline) &                          0.179&      1.408&      5.943&      0.247&      0.756&      0.920&      0.968\\
       Ours (sampling) &                          0.128&      0.844&      4.638&      0.192&      0.849&      0.958& \st{0.984}\\ 
       Ours (L2-smooth) &                         0.117&      0.786&      4.559& \ft{0.183}&      0.868& \st{0.962}& \ft{0.985}\\ 
       Ours ($w_d=1.0$) &                         0.230&      1.607&      6.775&      0.304&      0.624&      0.868&      0.953\\

       \red{Ours (P3P)} &                         \red{0.117}&      \red{0.789}&      \red{4.580}& \red{\st{0.184}}&      \red{0.870}&      \red{0.960}& \red{\ft{0.985}}\\
       \red{Ours (w/o PANet)} &                   \red{0.118}&      \red{0.795}&      \red{4.567}&      \red{0.186}&      \red{0.867}& \red{\st{0.962}}& \red{\ft{0.985}}\\

       Ours (w/o $\mathcal{L}_{rp}$) &            0.124&      0.788&      4.555&      0.187&      0.857&      0.960& \ft{0.985}\\
       Ours (w/o MED) &                           0.118&      0.798& \st{4.534}&      0.185&      0.866&      0.961& \ft{0.985}\\
       Ours (full) &                              0.118& \st{0.787}& \ft{4.488}& \ft{0.183}&      0.870& \st{0.962}& \ft{0.985}\\
       \hline
    \end{tabular}
    }
\end{table*}

\section{Experiments}
\textbf{Dataset.} The KITTI dataset \cite{Geiger-ijrr13} provides videos of 200
street scenes captured by RGB cameras, with sparse depth ground truths captured by laser scanner. 
For depth and flow evaluation, training was done on the KITTI raw \cite{Bian-nips2019,Zhao-cvpr20,Godard-iccv19}
and the frames were resized to $832\times256$ pixels. The depth was evaluated on the Eigen's testing split \cite{Eigen-nips14}, 
and the optical flow on the KITTI 2015 training set. 
For KITTI odometry evaluation, the standard setting \cite{Bian-nips2019,Zhao-cvpr20,Godard-iccv19}, 
which uses sequences 00-08 for training and 09-10 for testing, was followed. 

The KAIST urban dataset \cite{jeong-ijrr19} 
has longer mileage and more complex traffic environments 
The sequences containing stereo images are \textit{urban18-39}, 
and we used \textit{urban26-39} to build the dataset, dropping the simple highway sequences \textit{urban18-25}.
For VO evaluation, we used \textit{urban26}, \textit{29}, \textit{30}, \textit{32}, \textit{34}, \textit{37},
\textit{38} for training and the others for testing.

\textbf{Implementation details.} The Adam \cite{Kingma-iclr15} optimizer was used, the learning rate was set to $10^{-4}$, and the
batch size to 10. The training epochs from the first to the last stages were 30, 30, and 20. 
At the first two training stages, the learning rate was decreased to $5\times10^{-5}$ in the last 10 epochs.
The LM iteration was set to 30 in training and inference. 
The coefficients $[\gamma_{f}, \gamma_{fs}, \gamma_{drp}, \gamma_{ds}, \gamma_{dc}, \gamma_{dba}, \gamma_c]$ were set to 
$[1, 0.1, 0.2, 10^{-3}, 0.5, 0.5, 0.002]$.

\textbf{Metrics.} The existing metrics of depth, optical flow and odometry were used for evaluation,
as in previous methods \cite{Godard-iccv19,Luo-pami20}. 


\begin{figure}[ht]
    \centering
    \includegraphics[width=10cm]{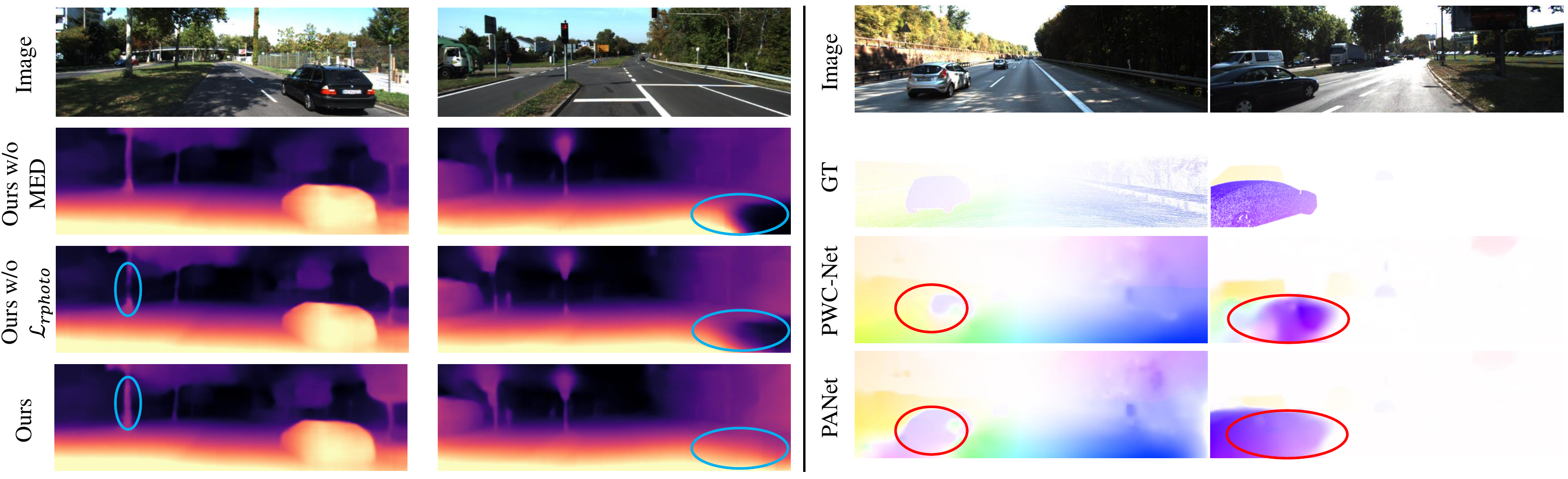}
    \caption{\red{Qualitative depth (left part) and optical flow (right part) results on the KITTI dataset.} 
    }
    \label{fig:depth_flow_ablation}
\end{figure}



\subsection{Depth evaluation}
\textbf{Ablation Study.} 
This part presents the ablation analysis used to assess the performance of the main components in our method. 
The definition of the metric can be found in \cite{Luo-pami20}. 
Several control experiments were conducted for evaluation (Table \ref{table:depth}), including:

(1) Ours (baseline): Depth-Net was trained with normal photometric error, edge-aware smoothness $\mathcal{L}_{s}$ and 
point-wise depth loss $\mathcal{L}_{ba}$, without using the inlier score $S_r$, and the mask $M_a$ to filter out the outlier. 

(2) Ours (sampling): based on baseline model, complete sampling strategies were added, which brings a great performance boost. 

(3) Ours (L2-smooth): training the model with all the components and using L2 smoothness loss $\mathcal{L}_{s}(\mathbf{d}, 1, 2)$ 
instead of L1 smoothness loss $\mathcal{L}_{s}(\mathbf{d}, 1, 1)$. Compared with L1 version, L2 has small performance degradation.

(4) Ours ($w_d=1.0$): Depth-Net was trained with all the components with $w_d=1.0$ for the depth Jacobi matrix in the training phase
\red{(Sec. \ref{sec:Method-BA} and Algorithm \ref{alg:inference})}.
It can be seen that this small weight is important when embedding BA modules into a deep learning framework in self-supervised learning.
Such a result may be due to the fact that changing $w_d$ means 
that optimization of the LM algorithm for depth and ego-motion 
makes the loss $\mathcal{L}_{rp}$ and $\mathcal{L}_{ba}$ conflict in training Depth-Net,
and thus falls into local minima. 

\red{
    (5) Ours (P3P): Depth-Net was trained using P3P algorithm with RANSAC \cite{gao-vslam} as pose estimation module. 
    Since P3P cannot update the depth, geometric BA and P3P was used to estimate the depth and ego-motion, respectively.
    As shown in Table \ref{table:depth}, compared with geometric BA, the P3P algorithm introduces a bit performance degradation.
}

\red{
    (6) Ours (w/o PANet): Depth-Net was trained using PWC-Net instead of proposed PANet as optical flow prediction module.
    The results in Table \ref{table:depth} shows that PANet is outperforms PWC-Net in our depth model.
}

(7) Ours (w/o $\mathcal{L}_{rp}$): Depth-Net was trained with all the components without the normalized photometric error $\mathcal{L}_{rp}$. 
Fig. \ref{fig:depth_flow_ablation} shows the qualitative results of the depth prediction using the proposed method. 
It is clear that the normalized photometric error $\mathcal{L}_{rp}$ shows advantages in the dark regions.

(8) Ours (w/o MED): Depth-Net was trained without MED. We can see that MED is better than Sigmoid function as the output activation process.

\textbf{Comparison with state-of-the-art methods.} 
The predicted depths are reported and compared with several methods in Table \ref{table:depth}, which indicates that
the proposed model achieves comparable performance with the methods \cite{Zhao-cvpr20,Godard-iccv19}.
Note that Depth-Net cannot converge without the loss $\mathcal{L}_{ba}$.
This is because using photometric error alone to train the network relies on the accuracy of the ego-motion estimation,
which is difficult to obtain from the BA module with few iterations and poor initial depth.
In addition, the absolute depth accuracy of our model is better than other methods in some areas (the second and fourth columns).
This is mainly due that BA loss $\mathcal{L}_{ba}$ can obtain relatively accurate depth at the later stages of training.

\begin{table}[t]
    \footnotesize
    \setstretch{0.8}
    \begin{minipage}[t]{0.5\textwidth}
        \centering
    \caption{Quantitative comparison of the proposed method with existing methods. 
    Bolded and underlined numbers is best and second-best metrics.}
    \label{table:flow}
    \resizebox{\linewidth}{!}{
    \begin{tabular}{lcccc}
        \hline
                & KITTI 2012 & \multicolumn{3}{c}{KITTI 2015} \\
        \cmidrule(r){2-2} \cmidrule(r){3-5}
        Method  & All$\downarrow$ & Noc$\downarrow$ & All$\downarrow$ & Fl$\downarrow$ \\
        \hline
        UnFlow \cite{Meister-aaai18}                & 3.29      & -         & 8.10       & 23.27\%      \\
        Geonet \cite{Yin-cvpr18}                    & -         & 8.05      & 10.81      & -            \\
        DF-Net \cite{Zou-eccv18}                    & 3.54      & -         & 8.98       & 26.01\%      \\
        CC \cite{Ranjan-cvpr19}                     & -         & -         & \st{5.66}  & 20.93\%      \\
        EPC++ \cite{Luo-pami20}                     & \st{2.30} & 3.84      & 5.84       & -            \\
        Zhao et al. \cite{Zhao-cvpr20}              & -         & \st{3.60} & 5.72       & \st{18.08\%} \\
        PANet (full)                                & \ft{1.83} & \ft{2.97} & \ft{5.23}  & \ft{16.52\%} \\
        \hline
        EPC++ \cite{Luo-pami20} (flow only)         & -         & \st{3.66} & \st{7.07}  & -            \\
        Zhao et al. \cite{Zhao-cvpr20} (flow only)  & -         & 4.96      & 8.97       & 25.84\%      \\
        Ours w/o PANet (flow only)                  & \st{3.53} & 3.74      & 8.64       & \st{22.49\%} \\
        PANet (flow only)                           & \ft{2.54} & \ft{3.57} & \ft{6.74}  & \ft{19.87\%} \\
        \hline
    \end{tabular}}
    \end{minipage} \quad
    \begin{minipage}[t]{0.5\textwidth}
        \centering
    \caption{Quantitative comparison of the proposed method with existing
    methods on the KITTI \cite{Geiger-ijrr13} VO dataset. 
    Bolded and underlined numbers denote respectively the best and second-best metrics 
    expect geometric algorithms DSO \cite{Engel-pami18} and ORB-SLAM \cite{Mur-tro17}.}
    \label{table:vo}
    \resizebox{\linewidth}{!}{
    \begin{tabular}{l*2{c}*2{c}}
       \hline
        & \multicolumn{2}{c}{Seq. 09} &\multicolumn{2}{c}{Seq. 10} \\
       Method & $t_{err}\downarrow$ & $r_{err}\downarrow$ & $t_{err}\downarrow$ & $r_{err}\downarrow$ \\
       \hline
       DSO \cite{Engel-pami18}            & 17.65     & 0.20      & 5.97      & 0.21      \\
       ORB-SLAM \cite{Mur-tro17}          & 9.31      & 0.26      & 2.66      & 0.39      \\
       \hline
       SfMLearner \cite{Zhou-cvpr17}      & 8.28      & 3.10      & 12.20     & 3.00      \\
       CC \cite{Ranjan-cvpr19}            & 6.92      & 1.80      & 7.97      & 3.10      \\
       SC-SfMLearner \cite{Bian-nips2019} & 7.31      & 3.05      & 15.04     & 6.34      \\
       Zhao et al. \cite{Zhao-cvpr20}     & 6.93      & 0.44      & 4.66      & 0.62      \\
       AdaptVO \cite{Li-cvpr20}           & 5.89      & 3.34      & 4.79      & 0.83      \\
       LongTermVO \cite{Zou-eccv20}       & 3.49      & 1.00      & 5.81      & 1.80      \\
       \hline
       Ours (baseline)                    & 5.46      & 1.54      & 4.74      & 1.98      \\
       Ours (sampling)                    & 5.26      & 1.71      & 4.41      & 2.05      \\
       Ours (baseline + BA)               & 3.35      & 0.47      & 4.07      & 1.54      \\
       Ours (sampling + BA)               & \ft{3.17} & 0.40      & 3.44      & 1.05      \\
       Ours (P3P init)                    & 3.57      & \st{0.30} & \st{2.97} & \st{0.32} \\
       Ours ($w_d=0.1$)                   & 4.15      & 1.02      & 4.71      & 1.33 \\
       \red{Ours (w/o PANet)}             &    \red{3.97} &    \red{0.35} &    \red{5.28} & \red{\ft{0.13}} \\
       Ours (full)                        & \st{3.32} & \ft{0.26} & \ft{2.96} & \st{0.29} \\
       \hline
    \end{tabular}}
    \end{minipage}
\end{table}


\subsection{Optical flow evaluation}
\label{sec:flow}
\textbf{Ablation Study.} The results of ablation study are presented in Table \ref{table:flow}. 
The metrics \textit{Noc} and \textit{All} are respectively the endpoint errors with non-occluded and all regions. 
\textit{Fl} indicates the percentage of erroneous pixels over all pixels. Three different variants of ablation study are reported, including:

(1) Ours w/o PANet (flow only): the model was trained with original PWC-Net \cite{Sun-cvpr18} and only trained at stage I 
with optical flow loss $\mathcal{L}_{flow}$. We can see that 
in the case of using PWC-Net \cite{Sun-cvpr18}, our method performs better than Zhao et al. \cite{Zhao-cvpr20} but somewhat worse than EPC++ \cite{Luo-pami20}.
This is probably because EPC++ \cite{Luo-pami20} does a better job of fine-tuning edge-aware smoothness loss $\mathcal{L}_{s}$.

(2) PANet (flow only): the model was trained with the proposed PANet and only trained at stage I. 
We can see that with the same training loss, PANet is much better than PWC-Net based Zhao et al. \cite{Zhao-cvpr20} and EPC++ \cite{Luo-pami20},
especially on the \textit{All} and \textit{Fl} for KITTI 2015. 

(3) PANet (full): the model was trained with the proposed PANet at all three stages. 
From the statistics, we notice that after the training at stage III (using cross-task loss $\mathcal{L}_{cross}$), 
the \textit{All} of PWC-Net based methods has significant improvement.

Fig. \ref{fig:depth_flow_ablation} shows that the proposed
PANet outperforms the optical flow model from PWC-Net with self-supervised training in some regions.

\textbf{Comparison with state-of-the-art methods.} Table \ref{table:flow} shows the quantitative results of 
self-supervised optical flow estimation on the KITTI 2015 training set. 
We can see that the proposed PANet outperforms other methods in self-supervised optical flow estimation under the same configuration. 
In addition, the optical flow module can benefit from cross-task loss $\mathcal{L}_{cross}$ for joint learning. 

\subsection{Odometry evaluation}
\textbf{Ablation study.}
The proposed model uses geometry-based algorithm to estimate odometry, 
thus, in this section we give an ablation analysis for VO inference.

(1) Ours (baseline): Sampling only uses flow occlusion score $S_o$ 
to filter out the outlier without inlier score $S_r$ and mask $M_a$.
We sampled 3000 points from the top 20\% of scoring regions.
After sampling the correspondence, camera motion inference was done using P3P algorithm without geometric BA. 

(2) Ours (sampling): The full sampling strategy is used on the basis of baseline version. 
Compared with the baseline, 
the complete sampling does not significantly improve the performance in the case of using P3P to estimate pose.

(3) Ours (baseline + BA): Based on the setting of \textit{Ours (baseline)}, geometric BA is used to infer the pose.
We set the initial rotation matrix to be a unit matrix, the initial translation vector to be zeros, 
and the initial damping factor $\lambda=0.1$, the same as in DSO \cite{Engel-pami18}.
The iteration number is set to 30.
Compared with P3P, BA performs better in our model.

(4) Ours (sampling + BA): The full sampling strategy is used on the basis of \textit{Ours (baseline + BA)} version. 
From this result, we can find that the complete sampling strategy improves the performance of pose estimation 
more than simple sampling (flow occlusion mask $M_c$) after using the BA algorithm.

(5) Ours (P3P init): Based on \textit{Ours (sampling + BA)} version, the method uses P3P as initialization.
It can be seen that using P3P gives better results than using fixed initialization, 
especially in terms of rotation error $r_{err}$ (1.05 vs. 0.32) for sequence 10.

(6) Ours ($w_d=0.1$): Based on \textit{Ours (full)} version, here we set $w_d=0.1$ instead of 1.0 in inference phase.
As we can see, contrary to the depth performance in the training phase (Table \ref{table:depth}), 
setting $w_d=0.1$ during inference leads to a decreased performance.

\red{
(7) Ours (w/o PANet): PWC-Net is used for optical flow estimation instead of the proposed PANet.
As we can see, relatively inaccurate optical flow estimation can directly affect the performance of VO.
}

(8) Ours (full): A dynamic damping factor $\lambda$ (see Equation \ref{eq:damp_factor}) is used 
instead of a fixed value 0.1 as initialization. 
As we can see, this component can further improve performance.

\begin{figure}[t]
    \centering
    \includegraphics[width=10cm]{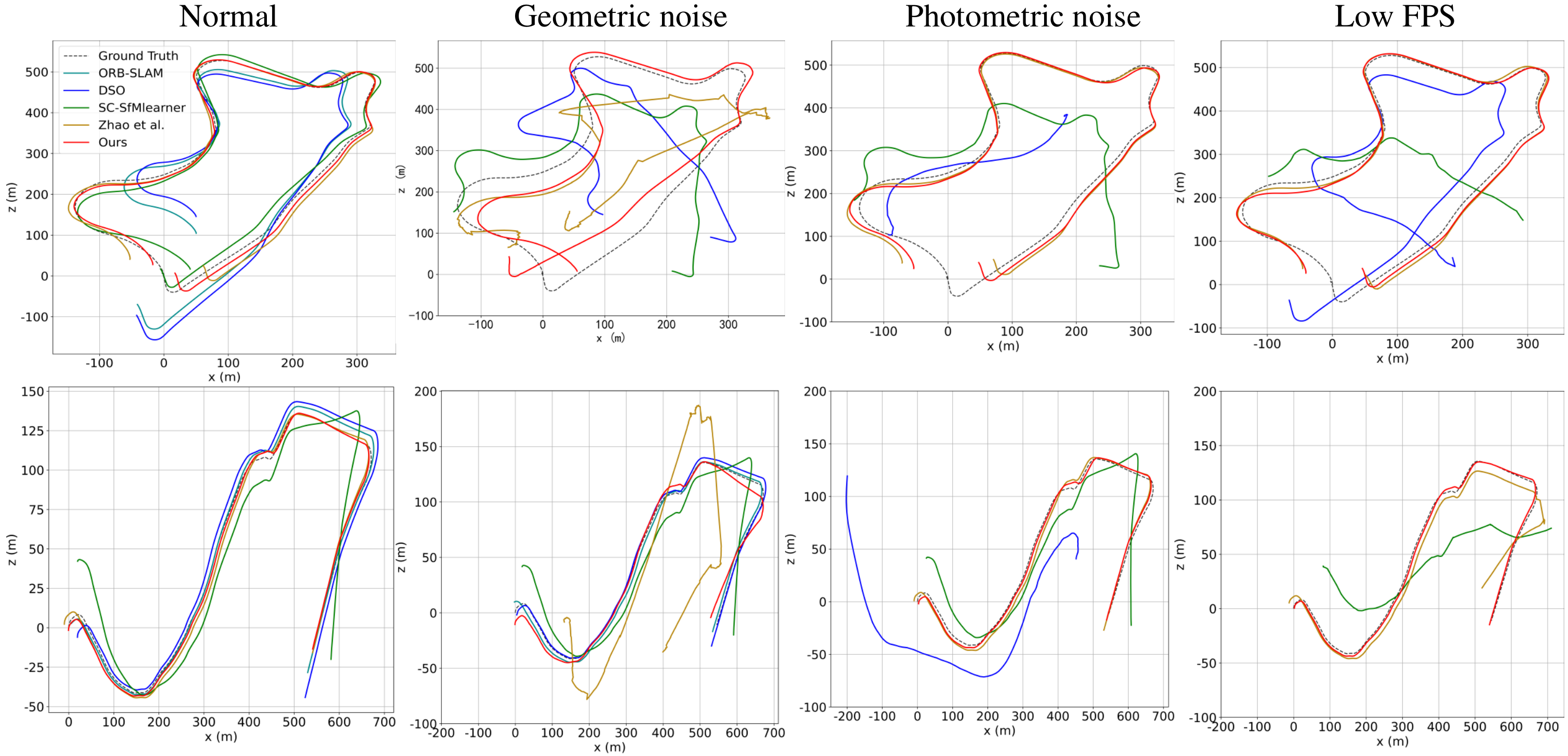}
    \caption{\red{Visualization of predicted trajectories by different models on the KITTI dataset: sequence 09 (top) and 
    sequence 10 (right).}}
    \label{fig:vo-normal-noise}
\end{figure}

\textbf{Comparison with state-of-the-art methods.} 
The trajectories of sequences 09-10 in the KITTI dataset are plotted in Fig. \ref{fig:vo-normal-noise} 
and quantitative results are shown in Table \ref{table:vo}. As can be seen, the proposed method significantly outperforms
all the other learning-based VO methods, 
including a long-term model trained with 97 frames \cite{Zou-eccv20} 
and the algorithms \cite{Li-cvpr20} trained online on a test set. 
Compared with a pure geometry-based algorithm, 
such as ORB-SLAM \cite{Mur-tro17} or DSO \cite{Engel-pami18}, 
the proposed model also achieves remarkable advantage in sequence 09.



\begin{table*}[t]
    \footnotesize
    \setstretch{0.8}
    \centering
    \caption{Quantitative VO results on the KITTI odometry dataset of synthetic images with
    geometric noise $\delta_g=3$, photometric noise $\delta_p=6$, and low FPS $\delta_s=3$.
    Symbol X indicates that the method failed to work under the corresponding conditions.}
    \label{table:robustVO}
    \resizebox{\textwidth}{!}{
    \begin{tabular}{l*4{c}*4{c}*4{c}}
        \hline
        & \multicolumn{4}{c}{Geometric noise}& \multicolumn{4}{c}{Photometric noise} & \multicolumn{4}{c}{Low FPS}\\
        & \multicolumn{2}{c}{Seq. 09} &\multicolumn{2}{c}{Seq. 10} &\multicolumn{2}{c}{Seq. 09} &\multicolumn{2}{c}{Seq. 10} &\multicolumn{2}{c}{Seq. 09} &\multicolumn{2}{c}{Seq. 10} \\
        Method & $t_{err}\downarrow$ & $r_{err}\downarrow$ & $t_{err}\downarrow$ & $r_{err}\downarrow$ & $t_{err}\downarrow$ & $r_{err}\downarrow$ & $t_{err}\downarrow$ & $r_{err}\downarrow$ & $t_{err}\downarrow$ & $r_{err}\downarrow$ & $t_{err}\downarrow$ & $r_{err}\downarrow$ \\
        \hline
        DSO \cite{Engel-pami18}            & 66.18      & 29.16     & 69.75     & 34.70     & 79.43     & 39.40     & 77.70     & 26.72     & 23.91     & 3.50      & X         & X     \\
        ORB-SLAM \cite{Mur-tro17}          & 10.55      & \ft{0.59} & 8.10      & \ft{1.30} & X         & X         & X         & X         & X         & X         & X         & X     \\
        SC-SfMLearner \cite{Bian-nips2019} & 28.47      & 6.27      & 14.60     & 6.55      & 34.89     & 7.71      & 16.95     & 8.39      & 55.57     & 14.78     & 35.12     & 17.72 \\
        Zhao et al. \cite{Zhao-cvpr20}     & 41.26      & 21.87     & 47.99     & 24.02     & \st{7.34} & \st{0.57} & \st{6.41} & \st{0.51} & 6.97      & \st{0.49} & 6.22      & 2.45  \\
        Ours (sampling)                    & \ft{5.63}  & \st{1.86} & \st{6.93} & 3.33      & 9.57      & 2.15      & 6.84      & 3.34      & 7.03      & 1.07      & \st{4.71}  & \st{0.97}  \\
        \red{Ours (w/o PANet)}             & \red{\st{5.95}}  & \red{1.87}      &  \red{7.56} & \red{4.06}      & \red{8.35}      & \red{2.35}      & \red{10.29}     & \red{4.26}      & \red{\ft{5.50}} & \red{2.01}      & \red{26.98}      & \red{15.65}      \\
        Ours                               &    {7.54}  & 1.87      & \ft{5.37} & \st{2.33} & \ft{6.45} & \ft{0.50} & \ft{4.61} & \ft{0.43} & \st{6.35} & \ft{0.45} & \ft{3.42}  & \ft{0.55}  \\
        \hline
    \end{tabular}}
\end{table*}




\begin{table*}[t]
    \footnotesize
    \setstretch{0.8}
    \centering
    \caption{Quantitative comparison of the proposed method with existing
    methods on the KAIST urban dataset \cite{jeong-ijrr19}. 
    The reported results are absolute trajectory error.}
    \label{table:kaist-vo-static}
    \resizebox{\textwidth}{!}{
    \begin{tabular}{lccccccc}
        \hline
        Method & Training data & \tabincell{c}{urban27\\5.4km} & \tabincell{c}{urban28\\11.47km} & \tabincell{c}{urban31\\11.4 km} & \tabincell{c}{urban33\\7.6 km} & \tabincell{c}{urban35\\3.2km} & \tabincell{c}{urban39\\11.06 km} \\
        \hline
        DSO \cite{Engel-pami18}              & -     & X          & X            & X           & X           & X          & X           \\
        ORB-SLAM \cite{Mur-tro17}            & -     & X          & X            & X           & X           & X          & X           \\
        SC-SfMLearner \cite{Bian-nips2019}   & KITTI & 312.86     & 412.80       & 958.54      & 480.39      & 276.81     & 358.29      \\
        Zhao et al. \cite{Zhao-cvpr20}       & KITTI & 137.64     & \st{186.23}  & 683.83      & X           & 122.33     & 443.09      \\
        Ours (sampling)                      & KITTI & 332.34     & 315.43       & 980.14      & 356.98      & 101.85     & 457.40      \\
        \red{Ours (w/o PANet)}               & KITTI & \red{157.07}     & \red{382.09}       & \red{454.31}      & \red{336.70}      & \red{131.56}     & \red{437.12}      \\
        Ours                                 & KITTI & \st{124.46}& \ft{100.68}  & \st{578.89} & \st{110.79} & \ft{73.02} & \st{284.34} \\
        Ours                                 & KAIST & \ft{98.32} & 212.08       & \ft{339.23} & \ft{93.58}  & \st{85.64} & \ft{143.82} \\
        \hline
    \end{tabular}}
\end{table*}

\subsection{Generalization capability}

\textbf{Odometry performance under noisy data.} 
Robustness is an important metric in a VO/SLAM system
to ensure that the system will not crash, or large errors will not occur in various unconventional scenes. 
In this work, two common cases of visual noise and low FPS were simulated using data synthesis; see Fig. \ref{fig:intro}.
The proposed method is compared with several representative algorithms, such as the direct method DSO \cite{Engel-pami18},
the indirect method ORB-SLAM \cite{Mur-tro17}, the learning-based VO SC-SfMLearner \cite{Bian-nips2019}, 
and a joint geometry-learning training model \cite{Zhao-cvpr20}. 


\red{
Geometric and photometric noise was synthesized in the same way as \cite{Engel-pami18}.
As for the low-FPS case, the original sequence was sampled separately with a large stride $\delta_s$, 
with $\delta_s = 3$. 
Table \ref{table:robustVO} shows the quantitative results with $\delta_g=3$, $\delta_p=6$, and $\delta_s=3$, respectively. 
}

According to the analysis in \cite{Engel-pami18}, ORB-SLAM is significantly more resistant to geometric noise, 
while DSO performs better in the presence of strong photometric noise.
For geometric noise, as expected, the learning-based SC-SfMLearner \cite{Bian-nips2019} shows poor performance due to the fact that 
geometric noise changes the distribution of the original image pair.
Unexpectedly, the hybrid learning- and geometry-based method \cite{Zhao-cvpr20} performs particularly poorly under geometric noise 
($t_{err}=41.26$ and $r_{err}=21.87$ in sequence 09 with $\delta_g=3$),
which is due to the complete failure of optical flow estimation of PWC-Net \cite{Sun-cvpr18} under this condition.

For photometric noise, ORB-SLAM often fails to track under this case, while DSO and SC-SfMLearner cannot give rewarding results. 
This is because feature-based ORB-SLAM fails to detect stable feature points, while 
DSO must ensure minimizing photometric error without requiring a relatively exact feature matching. 
Learning-based SC-SfMLearner cannot predict accurate ego-motion without being trained with corresponding noise data.
The joint geometry- and learning-based algorithm exhibits robustness under this noise condition. 
Zhao et al. \cite{Zhao-cvpr20} obtains a satisfactory prediction, while the proposed model further improves the performance.

Low-FPS sequences can simulate many situations, such as fast speeds of robot and vehicles, and low FPS caused by camera. 
Regarding geometry-based ORB-SLAM and DSO, the difficulty on such sequences is the lack of sufficient number of feature points to match,
which will cause the trace to fail. In this case, the hybrid learning- and geometry-based model outperforms other types of algorithms and exhibits good performance.

\red{Table \ref{table:robustVO} shows that the proposed model outperforms the model without geometric BA (Ours (sampling)) and 
the model inference with PWC-Net (Ours (w/o PANet)) in most of the noise experiments.
These results show that the proposed method is better in terms of stability when facing various types of noise.}

\textbf{Generalization on KAIST urban dataset.}
To verify the performance of the model in more realistic traffic scenarios, 
we employed the KAIST urban dataset \cite{jeong-ijrr19} to further evaluate the accuracy of VO.
This dataset has long mileage (3-12km), fast light changes, 
complex environments (including pedestrians, oncoming traffic and urban complexes) and motion closer to real driving.
Note that we only used absolute trajectory error (ATE) as the metric for evaluation, because 
the relative error $t_{err}$ and $r_{err}$ can not reflect the real situation 
in the long-range and large error trajectory evaluation.


Table \ref{table:kaist-vo-static} shows the quantitative experiments on VO estimation. 
To evaluate the generalization ability of our model, first, we directly used the model pre-trained on 
the KITTI dataset to test on the KAIST urban dataset \cite{jeong-ijrr19}.
As we can see, compared with pure learning method \cite{Bian-nips2019}, 
the geometric BA module introduced in this work makes the model more generalizable due that 
the model only learns low-level optical flow feature and is not sensitive to the overall distribution.
Pure geometry methods, such as DSO \cite{Engel-pami18} and ORB-SLAM \cite{Mur-tro17},
fail to track in the tested sequences. 
\red{The proposed model has a significant decrease in performances after removing PANet or geometric BA.}
In addition, the performance of the most sequences are further improved when the model was trained on the KAIST data.

\begin{figure}[t]
    \centering
    \includegraphics[width=6cm]{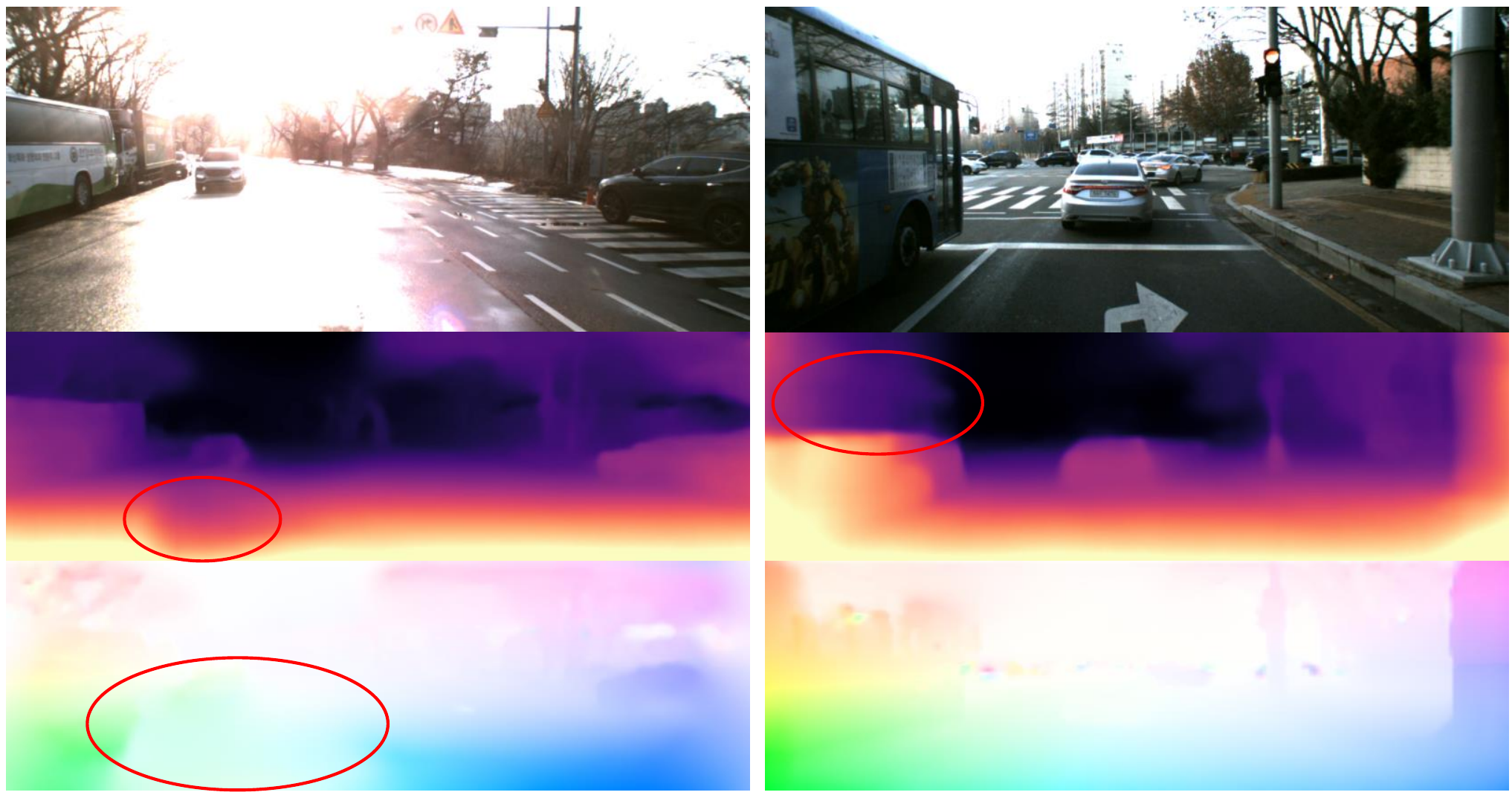}
    \caption{\red{Failure cases.}}
    \label{fig:failure_cases}
\end{figure}

\section{Failure cases}
\red{
Note that although the proposed optical flow and depth estimation modules have good generalizability, 
they do not perform well in some environments.
Figure \ref{fig:failure_cases} shows two typical failure cases.
In the high luminance environment, it is difficult to find a good local structure for monocular depth estimation 
and a stable matching for optical flow estimation, so it is difficult to get good results for both.
In some dynamic environments, with the objects like buses or black cars that are very close to the camera, 
the proposed monocular depth estimation algorithm usually has difficulty in estimating the accurate results.
}

\section{Discussion and conclusions}
\red{
VO performance mainly depends on 1) the accuracy of detecting and matching pixels among frames, and 2) follow-up VO estimates. 
Pure geometry-based algorithms lack sufficient fineness and robustness in pixel detection and matching, 
while a pure learning based model performs poorly on ego-motion estimate due to the lack of geometry restriction. 
Under more difficult conditions, these disadvantages will be magnified; as reflected in Table \ref{table:robustVO} and \ref{fig:vo-normal-noise}.
}

\red{
Therefore, learning depth and optical flow and predicting ego-motion using a
geometry-based approach is perhaps one of the feasible directions to solve the generalizability problem. 
In the proposed model, the gain on robustness comes from two designs: 
1) accurate and robust optical flow, and 
2) the BA module.
With the improvements made in the proposed method, the geometric BA module can be directly embedded into the joint learning
models to achieve remarkable depth, optical flow, and VO performance.
}

\red{
Joint geometry-learning has more advantages in generalization ability than the pure learning algorithms, 
as shown in Table \ref{table:robustVO}. 
This may be due that pure learning algorithms need to learn high-level information, such as ego-motion, 
while joint learning algorithms only need to learn the pixel-level image features (optical flow and depth), 
which vary relatively little across datasets.
}

\red{To conclude, this paper proposed} a novel optical flow network named PANet, 
which significantly improves the performance on self-supervised learning,
and the robustness under the noise condition. 
For better estimation accuracy and ego-motion generalization, 
we embed the geometric BA module into a joint depth and optical flow learning system and train them together in a self-supervised manner.
\red{To improve the performance of the proposed model and reduce the failure cases for complicated environments, 
in the future we plan to design a network that can estimate both optical flow and its uncertainty, 
hoping to obtain more stable correspondence based on uncertain estimation.}



\section*{Acknowledgment}
This work was supported by National Science and Technology Innovation 2030 Major Program (2022ZD0204600) 
and National Natural Science Foundation of China (62076055).

\bibliographystyle{elsarticle-num} 
\bibliography{ms}

\end{document}